\documentclass{egpubl}
\usepackage{eg2019}

\ShortPresentation      
 \electronicVersion 

\ifpdf \usepackage[pdftex]{graphicx} \pdfcompresslevel=9
\else \usepackage[dvips]{graphicx} \fi

\PrintedOrElectronic

\usepackage{t1enc,dfadobe}

\usepackage{egweblnk}
\usepackage{cite}
\usepackage{amsmath}
\usepackage{rotating}
\usepackage{amsfonts}
\usepackage[labelfont=bf, textfont=it]{caption} 
\usepackage{textcomp}

\title[DT-FCN for Motion Capturee Segmentation]
      {Fine-Grained Semantic Segmentation of Motion Capture Data using Dilated Temporal Fully-Convolutional Networks}

\author[N. Cheema \& S. Hosseini et al.]
{\parbox{\textwidth}{\centering N. Cheema$^{1,2,3}$\orcid{0000-0003-1275-4080},
        S. Hosseini$^{1,2}$\thanks{First two authors contributed equally; email: noshaba.cheema@dfki.de}\orcid{0000-0002-4674-7973}, J. Sprenger$^{1,2}$\orcid{0000-0002-6810-2674}, E. Herrmann$^{1,2}$\orcid{0000-0003-1052-9883}, H. Du$^{1,2}$\orcid{0000-0002-5956-6637}, K. Fischer$^{1,2}$\orcid{0000-0001-6393-2268} and P. Slusallek$^{1,2}$\orcid{0000-0002-2189-2429}
        }
        \\
{\parbox{\textwidth}{\centering $^1$German Research Center for Artificial Intelligence (DFKI) Saarbr{\"u}cken, $^2$Saarland University, $^3$Max-Planck Institute for Informatics; Germany
       }
}
}

\newcommand\ColorSquare[1]{{\color{#1}\rule{2mm}{2mm}}}
\begin{document}
\definecolor{none}{rgb}{0,0,0}
\definecolor{beginRightStep}{rgb}{1,0,0}
\definecolor{beginLeftStep}{rgb}{0,1,0}
\definecolor{rightStep}{rgb}{0,0,1}
\definecolor{leftStep}{rgb}{1,1,0}
\definecolor{endRightStep}{rgb}{1,0,1}
\definecolor{endLeftStep}{rgb}{0,1,1}
\definecolor{turnRight}{rgb}{0.72, 0.29, 0.94}
\definecolor{reach}{rgb}{1.0, 0.66, 0.07}
\definecolor{retrieve}{rgb}{0.39, 0.25, 0.15}



\teaser{
\vspace{-20pt}
\setlength{\belowcaptionskip}{5pt}
  \includegraphics[width=0.9\linewidth]{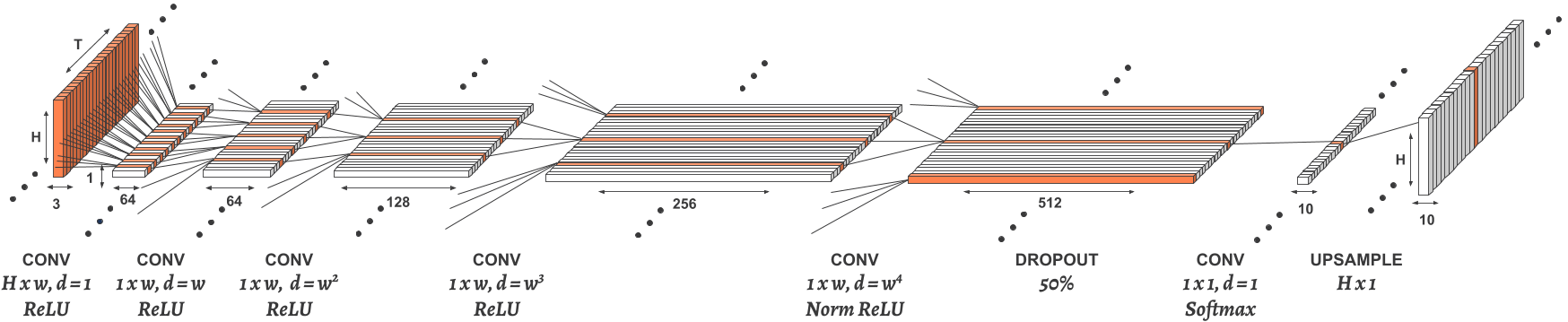}
  \centering
  \caption{Our dilated temporal fully-convolutional neural network (DT-FCN) for motion capture segmentation.}
    \label{fig:dtfcn}
}


\maketitle
\begin{abstract}
Human motion capture data has been widely used in data-driven character animation. In order to generate realistic, natural-looking motions, most data-driven approaches require considerable efforts of pre-processing, including motion segmentation and annotation. 
Existing (semi-) automatic solutions either require hand-crafted features for motion segmentation or do not produce the semantic annotations required for motion synthesis and building large-scale motion databases. In addition, human labeled annotation data suffers from inter- and intra-labeler inconsistencies by design. 
We propose a semi-automatic framework for semantic segmentation of motion capture data based on supervised machine learning techniques.
It first transforms a motion capture sequence into a ``motion image'' and applies a convolutional neural network for image segmentation. 
Dilated temporal convolutions enable the extraction of temporal information from a large receptive field. 
Our model outperforms two state-of-the-art models for action segmentation, as well as a popular network for sequence modeling. 
Most of all, our method is very robust under noisy and inaccurate training labels and thus can handle human errors during the labeling process.


\end{abstract}  
\section{Introduction}
Data-driven motion synthesis for digital human models has been widely used to generate realistic and natural human motion. Many data-driven techniques require motion segmentation as a necessary pre-processing step, for instance, statistical modeling \cite{min2012motion} and graph-based approaches \cite{kovar2008motion}. For semantic-embedded motion synthesis \cite{min2012motion,du2016joint}, recorded motions need to be split in structurally- and semantically-similar segments. An action can then be represented as a finite set of semantic-embedded states (motion primitives). For instance, walking can be decomposed as a combination of left and right steps. To our knowledge there is not extensive work done in motion primitive segmentation (i.e. segmenting \emph{left step} from \emph{right step} for example) using recognition-based segmentation methods, therefore most of the mentioned related work focuses on action segmentation (i.e. segmenting \emph{walking} from \emph{picking}). We include work based on 3D motion capture data, as well as, 2D video based data.

\textbf{Kinematic-based segmentation} methods such as \cite{muller2005efficient, min2012motion} commonly compare hand-crafted, low-level kinematic time-series features (e.g. distance foot to floor \cite{holden2016deep, min2012motion}) to find segment boundaries for such motion primitives. Since such features have to be designed separately for each new task, these methods are unfit for segmenting a multitude of different actions. I.e. the distance from foot to floor works for segmenting walking actions but to segment a picking action.

\textbf{Data-analysis-based segmentation} methods \cite{zhou2013hierarchical, vogele2014efficient} make use of unsupervised methods to automatically learn high-level features for segmentation. They have been the conventional method for segmenting skeletal motion data. Although, these approaches produce more sophisticated results and can be used for unseen data, they generally lack control of the feature selection and are not able to produce semantic labels as they cannot make any insights on the content or semantics of the motion data.

\textbf{Recognition-based segmentation}, on the other hand, is based on supervised learning approaches. Typically, in supervised motion segmentation techniques a collection of motion capture data is manually segmented and labeled to train a classifier. Therefore, segments created by the classifier can be as complex as segments created by humans. In recent years, deep learning methods have gained popularity in recognition-based segmentation. Recurrent neural networks (RNN) have become the go-to method to model time-dependent sequences and have also been used for action segmentation using motion capture data \cite{du2015hierarchical}. However, one of their major drawbacks is the exploding and vanishing gradient problem and the difficulty to parallelize their training. Recent studies \cite{bai2018empirical} suggest that certain architectures of Convolutional Neural Networks (CNNs) called Temporal Convolutional Networks (TCNs) can reach state-of-the-art results in typical sequence modeling tasks outperforming different types of RNNs. With the Copy Memory Task \cite{bai2018empirical}, they show that TCNs exhibit longer memory than recurrent architectures with the same capacity. TCNs have also been used for action segmentation in videos \cite{Lea2017TemporalCN}. The main advantage of recognition-based methods is the enhanced controllability; e.g. unsupervised methods like \cite{zhou2013hierarchical} segment a walking sequence to \emph{leftForward}, \emph{leftStep}, \emph{rightForward} and \emph{rightStop} - essentially, dividing a single step into half when using four clusters. Many motion synthesis approaches \cite{min2012motion} however, want to distinguish between a beginning/ending step and a step during locomotion. We therefore present a supervised segmentation method which is able to learn such semantic motion primitive labels. To account for inter- and intra-annotator disagreements, we further introduce noise into our training labels, as well as mask certain regions out. The presented method outperforms other state-of-the-art and popular recognition-based action segmentation methods.
\vspace{-5pt}

\section{Our Approach}
\begin{figure}
\setlength{\belowcaptionskip}{-15pt}
    \centering
    \includegraphics[width=0.9\linewidth]{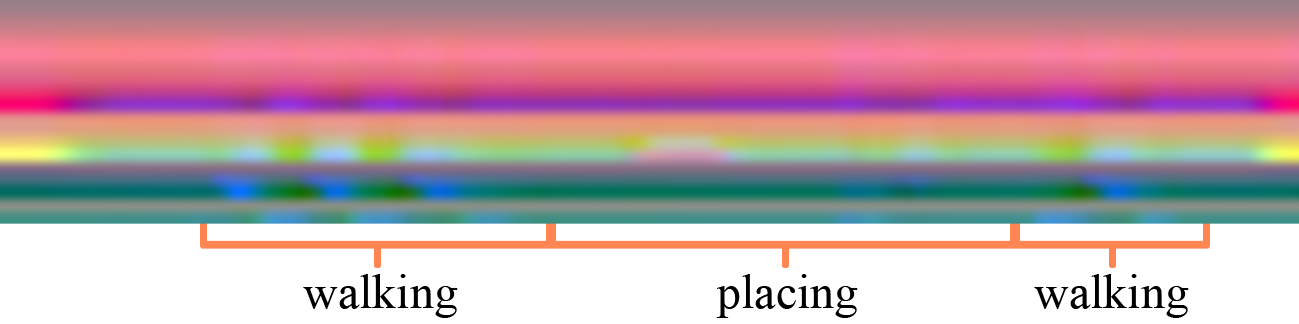}
    \caption{Sample generated motion image from local XYZ joint position coordinates.}
    \label{fig:motion_image_mixed}
\end{figure}

In a preprocessing step, we first transform our motion capture data into an RGB image, much in the spirit of \cite{laraba20173d}. Each column of the image represents a frame in the motion sequence. 
The rows represent the joints and the RGB values are the scaled XYZ Euclidean coordinates of each corresponding joint. Such a \emph{motion image} can be seen in Fig. \ref{fig:motion_image_mixed}. We then pass it to our network. The network architecture can be seen in Fig. \ref{fig:dtfcn}. Akin to \cite{long2015fully} our model has a total of five convolutional layers. Each layer multiplies the number of filters by two starting from 64 to 512 filters. By setting the kernel height of a 2D convolution in the initial layer to the height of the input image, it performs convolutions only in time domain and is able to process RGB image data. The next four layers are 1D temporal acausal convolutions with dilation. Every layer has the same convolution width $w$ with stride 1. Since motion features can span many frames, the filter needs to be able to look far into the future and into the past. Thus, our network requires a large receptive field. The receptive field of normal CNNs increase linearly to the depth of the network, thus increasing the amount of layers and parameters to train. Following the work of \cite{wavenet, bai2018empirical}, we apply dilated convolutions that enable exponentially large receptive field sizes. The dilation allows the filter to operate on a coarser scale than a normal convolution. For 1D sequence $\textbf{x} \in \mathbb{R}^N$ and filter $f: \{0, ..., k-1\} \rightarrow \mathbb{R}$ such a convolution is formally defined as 
$
(\textbf{x} *_d f)(t) = \sum_{i=0}^{k-1} f(i) \cdot \textbf{x}_{s-d \cdot i}
$
where $d$ is the dilation and $k$ the filter length. A normal convolution thus can be seen as a special case of a dilated convolution with $d=1$. 
Previous work \cite{bai2018empirical, Lea2017TemporalCN, wavenet} suggests that the dilation rate should be set to $d = 2^{l-1}$. However, we find that a rate of $d = w^{l-1}$ (layer number $l = 1 \dots L$, filter width $w$) is sufficient enough to ensure that there is at least one parameter hitting each input within the network's receptive field. Due to the increased dilation rate for $w>2$ a sufficient receptive field size can be achieved with less layers. To ensure that the values created by the ReLU activation layers before the Softmax function, do not exceed reasonable input values, we normalize the output of the last ReLU activation using the following function:
$
NormReLU(x) = \frac{ReLU(x)}{\max(ReLU(x)) + \epsilon}
$ with $\max(x)$ being the maximum values of the input tensor $x$. 
We use a value of $\epsilon = 1e^{-5}$ and found that this greatly improves accuracy. Finally, we upsample the output to the full image height for visualization purposes. Since our model is fully-convolutional \cite{long2015fully} it is able to handle input sequences of variable length.

\section{Evaluation}
Our model's parameters are learned using the categorical cross entropy loss with Stochastic Gradient Descent and the Adam optimizer. We implemented the network using Keras with a Tensorflow back-end. In all of our experiments, we use non-randomized 7-fold cross-validation and 100 training epochs. Our motion capture data set for evaluation consists of 70 sequences in total, of which 60 are used for training and 10 for testing. The data set contains standing, walking, picking, placing, and turning actions. 19 joints were used. Table \ref{tab:dataset} shows the 10 motion primitive labels used for segmentation and their total number of frames in the data set. The sequences are between 13-22 seconds long at 72 FPS, captured with OptiTrack\texttrademark. The segmentation was done by human experts.

\begin{figure}
\centering
\includegraphics[width=\linewidth, height=5cm]{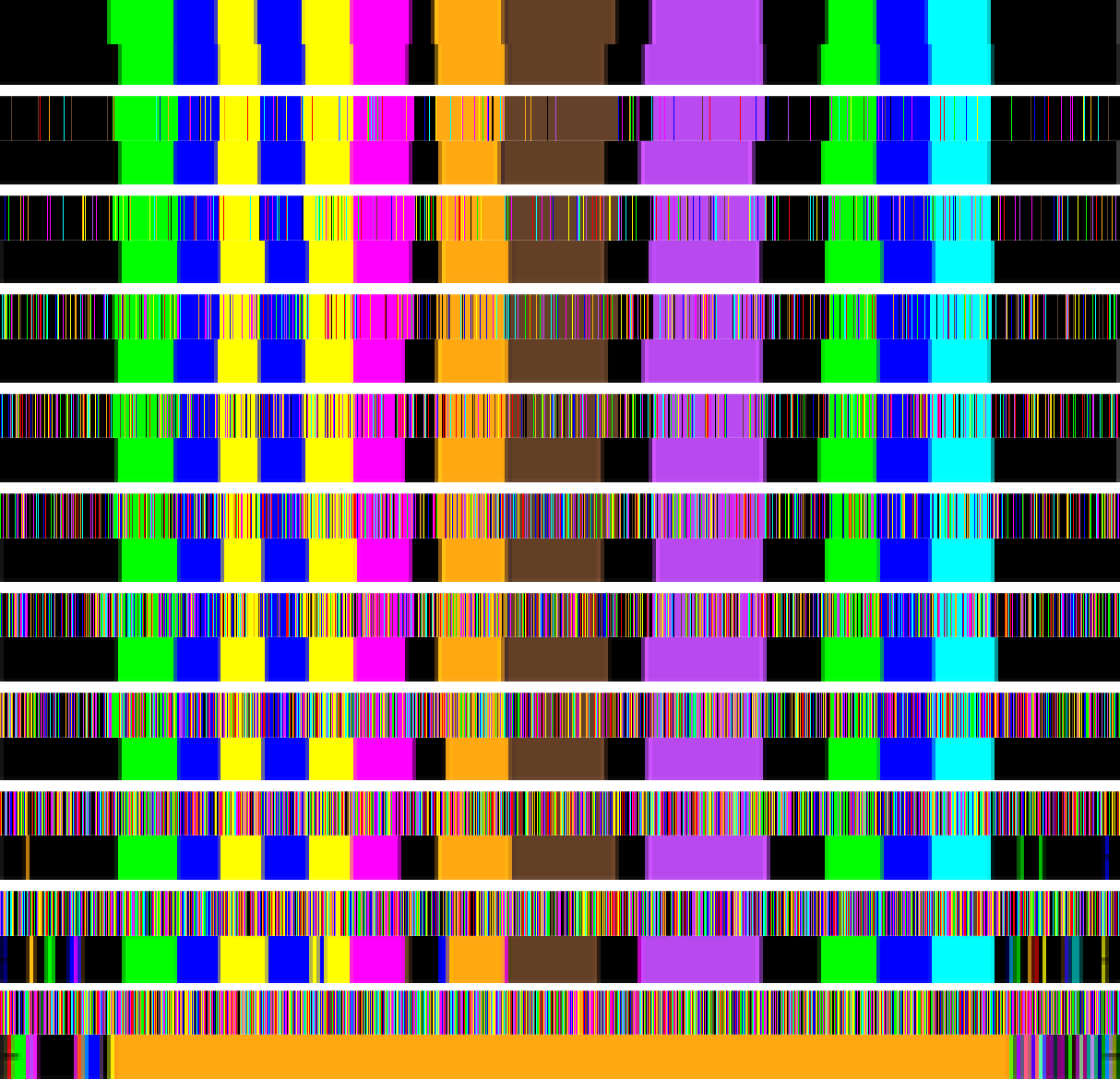}
\caption{Segmentation results of example training sequence of our $w = 3$ model. Noisy training label is on top of corresponding prediction. Top row without any noise. Noise added in +10\% intervals.}
\label{fig:our_noisy_pred}
\end{figure}

In order to determine the optimal receptive field size (RFS), we test our Dilated Temporal Fully-Convolutional Network (DT-FCN) on different convolution kernel widths $w$. Fig. \ref{fig:noise} shows  that even though a width $w = 5$ (RFS: 3125 frames) covers the longest sequence in the dataset, the test accuracy is worse on all noise levels compared to using $w = 3$ (RFS: 243 frames). This is in contrast to popular claims \cite{bai2018empirical, wavenet} in sequence modeling tasks that the receptive field of the network should at least cover the longest sequence in the data set, and supports a more intuitive suggestion that local structures are enough to be able to classify important structural and temporal information. Furthermore, the increase in parameter space for $w = 5$ make it also more prone to over-fitting, as compared to $w = 3$. Since our model has to be robust against human error due to wrongly-classified labels, we further train our model on noisy labels by setting a certain percentage of labels to a random label and test it on the true labels. Fig. \ref{fig:our_noisy_pred} shows such noisy labels for the training. Fig. \ref{fig:noise} shows despite adding 80\% noisy labels in the training data, an accuracy of almost 90\% is reached on the true test labels for $w = 3$. Qualitative results for $w = 3$ can be seen in Fig. \ref{fig:our_noisy_pred}.
We compare our best model ($w = 3$) against two other state-of-the-art TCN-based models \cite{lea2016temporal, Lea2017TemporalCN} for action segmentation and a commonly used RNN \cite{graves2005framewise} for sequence modeling using our data set without noisy labels. All TCN-based methods are able to train magnitudes faster compared to the RNN-based method, due to the ``embarrassingly parallel" nature of convolutions. Training takes $\sim$1 minute for the TCN-based methods compared to $\sim$40 minutes for Bi-LSTM for 100 epochs on a 4GB GTX970 and 16GB Intel-i7. Segmentation takes less than $\sim$1 second for all methods. Qualitatively, our model shows a better performance than the state-of-the-art (Tab. \ref{tab:results}, \ref{tab:noise_results}, Fig. \ref{fig:noise_bench_90}).

Discrepancies between human raters are mostly in the boundary regions between two adjacent segments. Hence, we conduct an experiment where a window of width $w_{transition}$ at a boundary frame contains randomly set, and hence wrong, training labels, as seen in Fig. \ref{fig:trans_noise_vs_no_inf} (top of each row). Additionally, we conduct an experiment which is less ``harsh" in nature: Instead of giving wrong information at the boundaries, we provide no information by setting the loss to zero within these regions. Essentially, leaving the networks to their own interpretation. The results of both experiments can be seen in Tab. \ref{tab:trans_noise_vs_no_inf} and Fig. \ref{fig:trans_noise_vs_no_inf}. Expectedly, results are better without any information than with wrong information. Most interestingly, masking the boundary frames (e.g. 11 frames, Tab. \ref{tab:trans_noise_vs_no_inf}) improves the overall model performance  compared to the original data (Tab. \ref{tab:results}). This might be due to inter- and intra-annotator disagreements: When there is no information given in these regions, the network is able to learn the relevant features from the geometric information alone. On average, this can be more accurate on unseen examples than learning from ``human generated'' annotations, which tend to be different from annotator to annotator (or even from using the same annotator) in boundary regions.

\begin{table}
\setlength{\belowcaptionskip}{-15pt}
    \centering
    \begin{tabular}{l|c|c|c}
       Motion Primitive&Description&Color&\#Frames \\
       \hline
       standing&standing in T or I pose& \ColorSquare{none}&29990\\
       begin left step &left step from standing & \ColorSquare{beginLeftStep}&2173\\
       begin right step&right step from standing & \ColorSquare{beginRightStep}&6509\\
       left step&left step in locomotion&\ColorSquare{leftStep}&8958\\
       right step&right step in locomotion&\ColorSquare{rightStep}&7201\\
       end left step&left step to standing&\ColorSquare{endLeftStep}&4763\\
       end right step&right step to standing&\ColorSquare{endRightStep}&4797\\
       reach&reach out with arm&\ColorSquare{reach}&10659\\
       retrieve&retrieve with arm&\ColorSquare{retrieve}&6813\\
       turn (in standing)&turn body direction&\ColorSquare{turnRight}&6630\\
    \end{tabular}
    \caption{Motion primitive labels with their corresponding label color in the segmented images and number of frames in the dataset.}
    \label{tab:dataset}
\end{table}

\begin{figure}
\setlength{\belowcaptionskip}{-7pt}
\centering
\includegraphics[width=\linewidth]{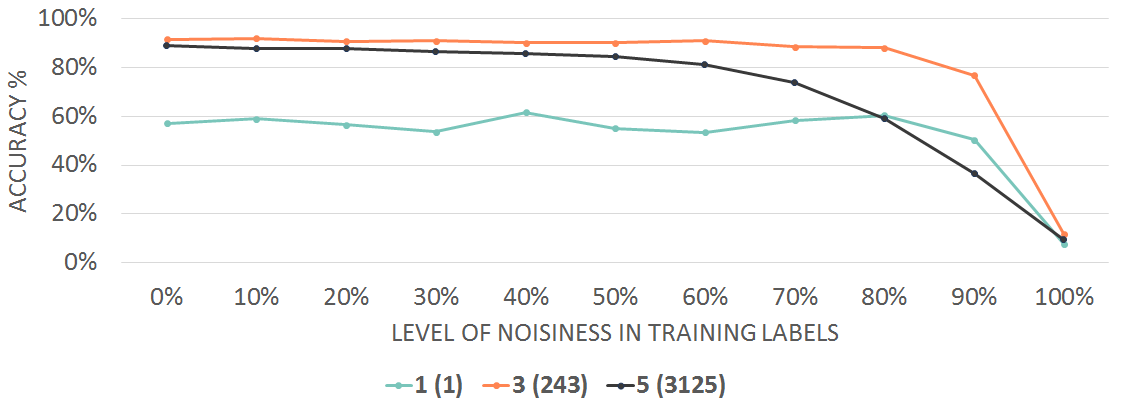}
\caption{Test accuracies for different receptive field sizes of our proposed architecture depending on training label noise level.}
\label{fig:noise}
\end{figure}

\begin{table}
\setlength{\belowcaptionskip}{-10pt}
\centering
\begin{tabular}{ |l|c|c|c|c| }
  \hline
  & Bi-LSTM & ED-TCN & DilatedTCN & Ours\\
   & \cite{graves2005framewise} & \cite{lea2016temporal} & \cite{Lea2017TemporalCN} & ($w = 3$)\\
  \hline
  Train & 86.32\% & 90.05\% & 90.64\% & \textbf{95.42\%}\\
  Test & 81.95\% & 88.69\% & 88.47\% & \textbf{90.81\%}\\
  \hline
  \# Param. & 378,634 & 1,613,770 & 388,874 & 663,562\\
  \hline
  RFS & $\infty$ & 49 & 254 & 243\\
  \hline
  Causality & Acausal & Acausal & Causal & Acausal\\
  \hline
\end{tabular}
\caption{Per-frame accuracy for benchmark models and our best model ($w = 3$) on the dataset without any noise in the train labels.}
\label{tab:results}
\end{table}

\begin{figure}
\setlength{\belowcaptionskip}{-7pt}
\includegraphics[width=\linewidth, height=2cm]{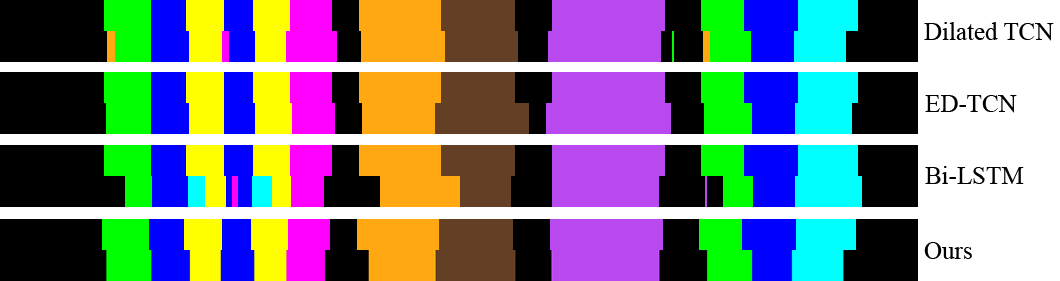}
\caption{Segmentation results of benchmark models, as well as our best model ($w = 3$) on an example test sequence. Each row shows ground truth (top half) and the corresponding prediction (bottom half).}
\label{fig:results00}
\end{figure}

\begin{table}
\centering
\scalebox{0.9}{
\begin{tabular}{ |c|c|c|c|c| }
  \hline
  Noise in & Bi-LSTM & ED-TCN & DilatedTCN & Ours\\
   Training & \cite{graves2005framewise} & \cite{lea2016temporal} & \cite{Lea2017TemporalCN} & ($w = 3$)\\
  \hline
  0\% & 81.95\% & 88.69\% & 88.47\% & \textbf{90.81\%}\\
  10\% & 83.70\% & 89.16\% & 89.37\% & \textbf{90.99\%}\\
  20\% & 79.94\% & 89.46\% & 88.86\% & \textbf{90.75\%}\\
  30\% & 85.02\% & 89.40\% & 88.90\% & \textbf{91.20\%}\\
  40\% & 84.29\% & 89.02\% & 88.64\% & \textbf{91.13\%}\\
  50\% & 81.72\% & 87.40\% & 87.93\% & \textbf{91.04\%}\\
  60\% & 83.08\% & 88.65\% & 86.14\% & \textbf{92.99\%}\\
  70\% & 82.16\% & 87.66\% & 79.46\% & \textbf{90.11\%}\\
  80\% & 81.91\% & 80.46\% & 61.82\% & \textbf{89.55\%}\\
  90\% & 71.74\% & 54.91\% & 34.99\% & \textbf{81.18\%}\\
  100\% & 8.76\% & 10.39\% & 11.85\% & 11.24\%\\
  \hline
\end{tabular}
}
\caption{Per-frame test accuracy for benchmark models and our best model ($w = 3$) with noise added into the training labels.}
\label{tab:noise_results}
\end{table}


\begin{figure}
\setlength{\belowcaptionskip}{-10pt}
\centering
\includegraphics[width=\linewidth]{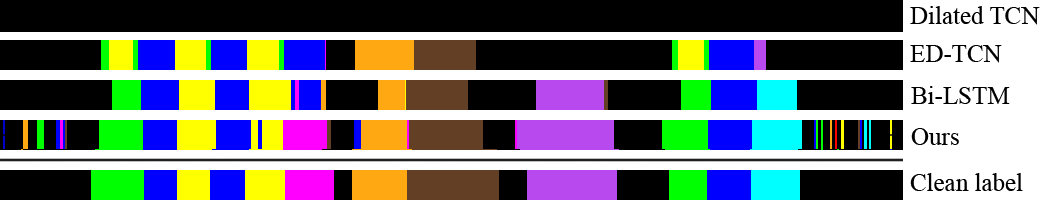}
\caption{Segmentation results for training on 90\% label noise.}
\label{fig:noise_bench_90}
\end{figure}

\begin{table}
\centering
\scalebox{0.9}{
\begin{tabular}{ |c|c|c|c|c| }
  \hline
   & Bi-LSTM & ED-TCN & DilatedTCN & Ours\\
   $w_{transition}$ & \cite{graves2005framewise} & \cite{lea2016temporal} & \cite{Lea2017TemporalCN} & ($w = 3$)\\
  \hline
  11 frames & 84.57\% & 88.15\% & 87.50\% & \textbf{91.45\%}\\
  21 frames & 81.47\% & 83.82\% & 85.48\% & \textbf{89.03\%}\\
  31 frames & 78.94\% & 75.75\% & 80.62\% & \textbf{83.59\%}\\
  \hline
  11 frames & 81.02\% & 88.25\% & 89.04\% & \textbf{92.42\%}\\
  21 frames & 80.05\% & 85.88\% & 86.80\% & \textbf{91.93\%}\\
  31 frames & 76.71\% & 84.88\% & 86.01\% & \textbf{90.48\%}\\
  \hline
\end{tabular}
}
\caption{Per-frame test accuracies for benchmark models and our best model ($w = 3$) with noise in the transition regions (top three rows) and with masked boundary regions (bottom three rows).}
\label{tab:trans_noise_vs_no_inf}
\end{table}


\begin{figure}
\centering
\setlength{\belowcaptionskip}{-15pt}
\includegraphics[width=\linewidth, height=2.5cm]{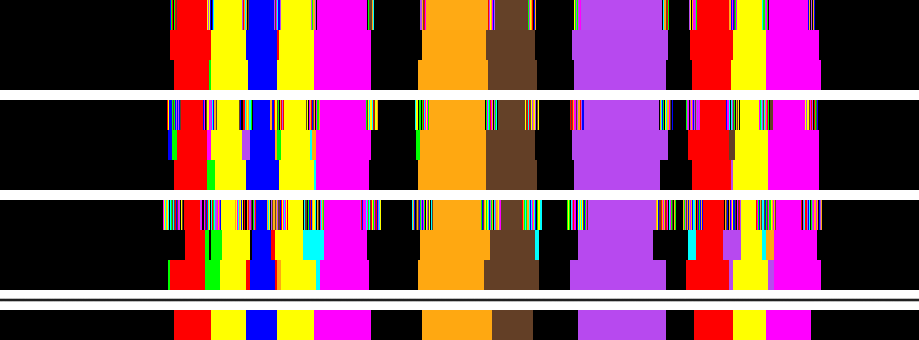}
\caption{Predicted segmentations by our model ($w = 3$). Blocks from top to bottom: $w_{transition} = 11$, $w_{transition} = 21$, $w_{transition} = 31$. Each block consists of the noisy training label (top), the corresponding prediction without masking out the noise (middle) and with masking out (bottom).}
\label{fig:trans_noise_vs_no_inf}
\end{figure}


\section{Conclusion}
In this paper we present a first dilated temporal FCN-based method for fine-grained semantic segmentation of motion capture sequences. Compared to commonly used unsupervised methods \cite{zhou2013hierarchical} our approach is able to learn complex labels such as "begin left step" or "end right step", while robustly handling labeling errors. While the key-ingredients for the success were already present in prior work \cite{laraba20173d, long2015fully, bai2018empirical}, we believe it is the combination of these methods which accounts for the improved accuracy compared to other TCN-based methods.
The combination of a VGG/FCN32-inspired model with acausal dilated convolutions and an increased dilation rate compared to other methods \cite{Lea2017TemporalCN, wavenet} enables a large receptive field with few parameters. Thus training time is reduced while the performance of the model is above state-of-the art competitors. It can even distinguish very similar motion primitives like \emph{start left step} and \emph{left step}.  Most of all, the model is very robust under label noise. Hence even cheaply labelled data can be used for training. Even more, we found that a semi-supervised approach by removal of the boundary labels between classes can improve the performance further. 
With this work, we have shown that our model provides a fruitful segmentation tool for motion capture segmentation.
\vspace{-5pt}

{\small
\section*{Acknowledgements}
This work is funded by the European Union's Horizon 2020 research and innovation program under the Marie Sklodowska-Curie grant agreement No 642841; and by the BMBF through the projects REACT (grant no. 01IW17003) and Hybr-iT (grant no. 01IS16026A), as well as the ITEA3 project MOSIM (grant no. 01IS18060C).}

\bibliographystyle{eg-alpha-doi}

\bibliography{sections/references.bib}
\end{document}